\documentclass[twoside]{article}

\usepackage[accepted]{aistats2025}

\usepackage[style=apa, natbib, backend=biber]{biblatex}
\renewcommand{\refname}{References}
\defbibheading{bibliography}{\subsubsection*{\refname}}
\AtEveryBibitem{\clearfield{publisher} \clearfield{pages} \clearfield{institution} \clearfield{location} \clearfield{isbn} \clearfield{url} \clearfield{month}\clearfield{language} \clearfield{issn}\clearfield{note} \clearfield{urldate}}
\usepackage{commands}

\begin{document}

%

%

\twocolumn[
    \aistatstitle{The VampPrior Mixture Model}
    \aistatsauthor{Andrew A. Stirn \And David A. Knowles}
    \aistatsaddress{Columbia University \And Columbia University \& New York Genome Center}
]

\begin{abstract}
Widely used deep latent variable models (DLVMs), in particular Variational Autoencoders (VAEs), employ overly simplistic priors on the latent space.
To achieve strong clustering performance, existing methods that replace the standard normal prior with a Gaussian mixture model (GMM) require defining the number of clusters to be close to the number of expected ground truth classes a-priori and are susceptible to poor initializations.
We leverage VampPrior concepts~\citep{tomczak_vae_2018} to fit a Bayesian GMM prior, resulting in the \emph{VampPrior Mixture Model} (VMM), a novel prior for DLVMs.
In a VAE, the VMM attains highly competitive clustering performance on benchmark datasets.
Integrating the VMM into scVI~\citep{lopez_deep_2018}, a popular scRNA-seq integration method, significantly improves its performance and automatically arranges cells into clusters with similar biological characteristics.
\end{abstract}

\section{INTRODUCTION}
\label{sec:introduction}

Extracting meaningful knowledge from complex, high-dimensional data is a central promise of data science.
A striking example is single-cell RNA sequencing (scRNA-seq), which has become ubiquitous in biomedical research, enabling genome-wide profiling of gene expression for millions of cells in one experiment.
However, single-cell datasets increasingly contain many samples collected under different conditions using different methodologies, leading to complex nested batch effects~\citep{luecken_benchmarking_2022} that often drive larger variance than the biological signals of interest.
The goal of scRNA-seq integration is to remove these batch effects while conserving biological variation such as cell type identity, disease, and perturbation effects.

The modus operandi of scRNA-seq analysis is to first integrate the observed data $X$, a $N$ cells $\times~G$ genes matrix, by mapping to an embedded space $Z$, with lower dimension $p \ll G$.
$Z$ is further reduced to visualizable dimensions with t-SNE~\citep{van_der_maaten_visualizing_2008}, UMAP~\citep{mcinnes_umap_2018}, or MDE~\citep{agrawal_minimum-distortion_2021}.
When the embedding function does not account for systematic shifts in expression profiling between datasets and/or batches that use different scRNA-seq technologies, misleading structure can arise, confounding standard analysis pipelines.
Accordingly, \citet{lahnemann_eleven_2020} identify atlas-level integration as one of the grand challenges of single-cell data science.

\citet{luecken_benchmarking_2022} benchmark 12 integration methods on 13 atlas-level integration tasks and find that Scanorama~\citep{hie_efficient_2019} as well as DLVMs using amortized variational inference (VI) such as scVI~\citep{lopez_deep_2018} and its derivatives~\citep{svensson_interpretable_2020,xu_probabilistic_2021} are the most performant, outperforming popular methods such as Harmony~\citep{xu_probabilistic_2021}.
scVI provides the encoder and decoder a one-hot encoded batch identifier to encourage a biological embedding space that is disentangled from confounding batch effects.
With a fixed $\N(0,I)$ prior on the embedding, scVI's generative process does not explicitly encourage clustering of similar cells.
We propose replacing scVI's $\N(0,I)$ prior with a more flexible mixture prior that both improves its integration performance and provides robust clustering capabilities.

To achieve simultaneous integration and clustering we first investigate a method, VaDE~\citep{jiang_variational_2017}, that replaces the $\N(0,I)$ prior of the standard Variational Autoencoder (VAE)~\citep{kingma_auto-encoding_2014, rezende_stochastic_2014} with a Gaussian mixture model (GMM).
We extend VaDE in two ways.
First, we reformulate the mixture as a Bayesian GMM with hyper-priors on component parameters and a Dirichlet hyper-prior on mixing proportions;
this allows us to instantiate an arbitrarily large number of components to closely approximate a Dirichlet process (DP) GMM~\citep{edward_infinite_2000} (i.e.\ a truncated DP) enabling our model to automatically prune unnecessary components.
Second, rather than learning point-estimates for cluster centers, we fit their distributions with the encoder network operating on trainable pseudo-inputs analogously to the VampPrior~\citep{tomczak_vae_2018}.
Initializing these pseudo-inputs with randomly selected training data ensures latent cluster centers are well initialized resolving VaDE's susceptibility to poor initializations.
We call our approach the \emph{VampPrior Mixture Model} (VMM), which we optimize using an algorithm that alternates between amortized VI steps (maximizing the variational objective with fixed prior parameters) and Empirical Bayes steps (fitting just prior parameters).

Since the VMM merely replaces a $\N(0,I)$ prior, it is applicable to any DLVM with continuous latent variables.
\Cref{sec:vae-experiments} employs a VMM within the VAE's generative process for image clustering.
The VMM not only outperforms VAE-based clustering methods but also approaches state-of-the-art unsupervised classification performance.
\Cref{sec:singe-cell} integrates the VMM into scVI and finds significant improvements to both batch correction and biological conservation during scRNA-seq integration, while also enabling unsupervised cell-type annotation.
\href{https://docs.scvi-tools.org/en/stable/user_guide/index.html}{Numerous other DLVMs} extending scVI have been developed for scATAC-seq, multimodal single-cell, and spatial transcriptomic data analyses that all use $\N(0,I)$ priors in their generative processes.
Integrating the VMM into these tools, among others, would be straightforward, amplifying this work's potential impact.

\section{BACKGROUND}
\label{sec:background}

\subsection{The Variational Autoencoder}
\label{subsec:vae}
DLVMs parameterize data generating distributions with neural networks operating on latent variables.
The VAE~\citep{kingma_auto-encoding_2014} is a DLVM with generative process,
\begin{align}
    \label{eq:vae-prior} z_i &\sim \N(0,I) \quad\forall~i\in[N]~ \\
    \label{eq:vae-likelihood} x_i | z_i &\sim p_{\theta}(x_i|z_i) \triangleq p(x_i|f(z_i;\theta)) \quad \forall~i\in[N],
\end{align}
where $f$ is the decoding neural network with parameters $\theta$.
The VAE employs amortized VI, defining the mean and covariance of the variational family $q_{\phi}(z_i;x_i)\triangleq \N(z_i;g(x_i;\phi))$ as outputs of neural network $g$ with parameters $\phi$ operating on observed $x_i$, and black-box VI~\citep{ranganath_black_2014} with reparameterization gradients~\citep{williams_simple_1992} to maximize the evidence lower bound (ELBO) \wrt $\theta$ and $\phi$,

\begin{small}
    \begin{align}
         \label{eq:vae-elbo} \LL(\D;\theta,\phi) = \sum_{x\in\D} & \E_{q_{\phi}(z;x)}\big[\log p_{\theta}(x|z) \big] - \dkl\big(q_{\phi}(z;x) || p(z)\big).
    \end{align}
\end{small}%

\subsection{The VampPrior}
\label{subsec:vampprior}

\citet{tomczak_vae_2018} identify the prior that maximizes \cref{eq:vae-elbo} as the aggregate posterior $p^*(z)=N^{-1}\sum_{i=1}^{N}q_{\phi}(z;x_i)$, over the $N$ training points.
To approximate the aggregate posterior, they replace the VAE's prior~\cref{eq:vae-prior} with their VampPrior,
\begin{align*}
    p(z) \triangleq \frac{1}{K}\sum_{j=1}^K q_{\phi}(z;u_j),
\end{align*}
where pseudo-inputs $u_1,\hdots,u_K$ are trainable prior parameters initialized with $K \ll N$ randomly selected training points for efficiency.
Inference proceeds by maximizing~\cref{eq:vae-elbo} \wrt $\theta$, $\phi$, and now $u$.

\subsection{Gaussian Mixture Models}
\label{subsec:gmm}

A GMM on $z$ can be represented as,
\begin{align}
    \label{eq:gmm} & z_1,\hdots,z_N | \pi, \mu, \Lambda \sim \sum_{j=1}^K \pi_j \N(z|\mu_j,\Lambda^{-1}_j),
\end{align}
with mixing proportions $\pi$, cluster means $\mu$, and cluster precisions $\Lambda$.
In a Bayesian framework, we place priors on all parameters,
\begin{align}
    \label{eq:concentration-prior}  \alpha &\sim \text{InverseGamma}(1,1), \\
    \label{eq:mixing-prior}  \pi|\alpha &\sim \text{Dirichlet}\big(\alpha K^{-1} \mathbf{1}_K^T\big), \\
    \label{eq:centers-prior-gmm}  \mu_k &\sim \N(0,I) & \forall ~ k \in [K], \\
    \label{eq:precisions-prior}  \Lambda_k &\sim \text{Wishart}\Bigg(p+2,\frac{K^{\frac{1}{p}}}{p+2} I\Bigg) & \forall ~ k \in [K],
\end{align}
where $p\equiv\dim(z)$.
Taking $K\rightarrow\infty$ results in a Dirichlet process (DP) GMM~\citep{edward_infinite_2000}.
\citet{ishwaran_approximate_2002} prove that an upper bound for the absolute difference between the marginal densities of a $K$-component GMM and the $\infty$-component GMM integrated over $\Re^p$ decays exponentially \wrt $K$.
Our Bayesian mixtures, which use large $K$, will therefore well approximate the limiting DP mixture and eliminate clusters in a similar manner.

\section{METHODS}
\label{sec:methods}
Before introducing the VMM, we first show how to use~\cref{eq:gmm}-\cref{eq:precisions-prior} as a hierarchical prior that replaces~\cref{eq:vae-prior} in the VAE\@.
We then describe our inference algorithm that alternates between VI and Empirical Bayes steps.
From there, the VMM is a straightforward change.

\subsection{A Bayesian GMM Prior}
\label{subsec:gmm-prior}

\citet{fraley_bayesian_2007}\('\)s recommended Wishart parameters for GMMs with observed $z_i$\('\)s motivated our choice in \cref{eq:precisions-prior}.
In our setting, $z_i$\('\)s are latent, so we replace the empirical precision term in the Wishart's scale matrix with the identity.
We found that further normalizing the scale matrix by $p+2$ to ensure $\E[\Lambda_j]=K^{\frac{1}{p}}I$ makes~\cref{eq:precisions-prior} more tolerant to different latent dimensions $p$.

For a VAE, defining $p(z)$ as a GMM changes only the Kullback-Leibler (KL) divergence term in~\cref{eq:vae-elbo} to,
\begin{align}
    \label{eq:dkl-gmm} \dkl\Bigg(q_{\phi}(z;x) \Big|\Big| \sum_{j=1}^K \pi_j \N(z|\mu_j,\Lambda^{-1}_j) \Bigg).
\end{align}

\subsection{Alternating Inference Algorithm}
\label{subsec:aternating-inference}

We use a Bayesian GMM prior to achieve a clustered latent representation without knowing the true number of classes a-priori.
In this scenario, we found jointly optimizing variational and prior parameters was suboptimal.
Instead, we partition all parameters into distinct variational ($\theta$ and $\phi$) and prior ($\alpha,\pi,\mu,\Lambda$) parameter sets and then alternate between variational inference and Empirical Bayes gradient steps\footnote{Direct optimization of $\theta$ can be viewed as part of VI by considering a uniform prior and defining $q(\theta)$ as a Dirac delta.}.
We perform VI (optimizing~\cref{eq:vae-elbo} \wrt $\theta$ and $\phi$) to fit variational parameters and perform maximum a posteriori (MAP) estimation (optimizing~\cref{eq:gmm}-\eqref{eq:precisions-prior} \wrt to $\psi\triangleq\{\alpha,\pi,\mu,\Lambda\}$) to fit prior parameters.

\begin{algorithm}[t]
    \caption{Alternating VI and Empirical Bayes}
    \label{alg:aternating-inference}
    \begin{algorithmic}
        \WHILE{not converged}
            \STATE Sample batch:
                $\B \leftarrow \{x_1,\hdots,x_M\sim\hat{\Pr}(\D)\}$
            \STATE VI:
                $(\phi,\theta) \leftarrow (\phi,\theta) + \gamma_1 \nabla_{\phi,\theta} \LL(\B;\theta,\phi)$
            \STATE Sample posterior: $z_i\sim q_{\phi}(z_i;x_i) \quad \forall~i\in[M]$
            \STATE EM: $\psi \leftarrow \psi + \gamma_2 \nabla_{\psi} \E_{q(c)}[\log p(z,c,\pi,\mu,\Lambda,\alpha)]$
	    \ENDWHILE
    \end{algorithmic}
\end{algorithm}

\Cref{alg:aternating-inference} shows how we alternate between these two separate inference procedures.
We first perform a step of stochastic amortized VI for the VAE as usual.
We then sample the aggregate posterior and perform a step of gradient based MAP estimation via expectation maximization (EM), learning point estimates for $\alpha,\pi,\mu,\Lambda$.
EM requires representing the GMM as,
\begin{align*}
    c_i | \pi \sim \text{Categorical}(\pi), \quad
    z_i | c_i, \mu, \Lambda \sim \N(z|\mu_{c_i},\Lambda^{-1}_{c_i}).
\end{align*}
The E-step is $q(c_i=j) \varpropto \log p(z_i|c_i=j, \mu, \Lambda) + \log \pi_j$.
We repeat this process until convergence, i.e.\ when validation set performance plateaus.

Cluster granularity and thereby performance depends on the strength of the hyper-priors in \cref{eq:concentration-prior,eq:mixing-prior,eq:centers-prior-gmm,eq:precisions-prior}.
We found that adjusting the batch size and/or the Empirical Bayes learning rate $\gamma_2$ was effective at regulating the strength of the hyper-priors and simpler than adjusting their many parameters.
A larger batch size up-weights the prior likelihood \cref{eq:gmm} relative to the hyper-priors during Empirical Bayes EM\@.

\subsection{The VMM}
\label{subsec:vmm}

The VMM retains the DP-GMM's generative process~\cref{eq:gmm}-\eqref{eq:precisions-prior}, but rather than learning point estimates for $\mu_1,\hdots,\mu_K$, it fits distributions over cluster centers $\mu_1,\hdots,\mu_K$ analogously to the VampPrior by defining their variational distributions as,
\begin{align}
    \label{eq:centers-vmm} \mu_j \sim q_{\phi}(\mu_j;u_j) \triangleq \N(\mu_j;g(u_j;\phi)) \quad \forall~j\in[K].
\end{align}
Thus, the standard normal prior on $\mu_1,\hdots,\mu_K$ \cref{eq:centers-prior-gmm} still regularizes cluster dispersion but indirectly via $u_1,\hdots,u_K$ rather than directly on $\mu_1,\hdots,\mu_K$.
Like the VampPrior, the VMM uses the encoder network operating on pseudo-inputs $g(u_j;\phi)$.
Where the VampPrior uses the encoder network to parameterize a mixture prior over the latent embedding $z$ directly, the VMM uses the encoder network to parameterize variational cluster distributions during Empirical Bayes.

Fitting distributions $q_{\phi}(\mu_j;u_j)$ instead of point estimates only requires substituting~\cref{eq:dkl-gmm} with,
\begin{align}
    \label{eq:dkl-vmm} \dkl\Bigg(q_{\phi}(z;x) \Big|\Big| \sum_{j=1}^K \pi_j \E_{q_{\phi}(\mu_j;u_j)}\big[\N(z|\mu_j,\Lambda^{-1}_j)\big] \Bigg)
\end{align}
during VAE inference, which we calculate analytically.
\Cref{eq:dkl-vmm} approaches that of the VampPrior for $\pi_j\rightarrow\frac{1}{K}$ and $\Lambda_j^{-1}\rightarrow 0$.

Following \citet{tomczak_vae_2018}, we initialize pseudo-inputs with randomly sampled training data.
After replacing $\mu$ with $u$ in parameter set $\psi$, \cref{alg:aternating-inference} adjusts easily to the VMM, requiring only that the E-step now also be \wrt $q_{\phi}(\mu;u)$ (the joint distribution for $\mu_1,\hdots,\mu_K$),
\begin{align*}
    \E_{q(c)q_{\phi}(\mu;u)}\big[\log p(z,c,\pi,\mu,\Lambda,\alpha)\big],
\end{align*}
which we also calculate analytically.
The M-step back propagates through $g(u_j;\phi)$, treating $\phi$ as a constant, to compute gradients \wrt $u_j$.
For a complete presentation of VMM updates, see \cref{sec:vmm-sup}.

We selected MAP EM for prior inference to preserve Gaussian amortized variational families and maintain compatibility with the many non-standard VAE-based methods used in scRNA-seq analysis~\citep{lopez_deep_2018,svensson_interpretable_2020,xu_probabilistic_2021}.
We seek only to replace the VAE's $\N(0,I)$ prior with a more flexible GMM prior (where the $\N(0,I)$ is now over cluster centers).
For example, had we chosen VI instead of MAP EM for prior inference, a Normal-Wishart $q(\mu_j,\Lambda_j)$ is required for analytic component distribution computation, $\E[\N(z|\mu_j,\Lambda^{-1}_j)]$, in~\cref{eq:dkl-vmm}.

Prior selection can be seen as a spectrum with the unimodal $\N(0,I)$ prior at one extreme and the exact VampPrior (i.e.\ the aggregate posterior) with a component for each training data point at the other.
The VMM falls between these extremes, identifying an appropriate number of modes to accurately model the data.
In the limit of a single cluster, the VMM can approach an unimodal $\N(0,I)$ prior.
In the limit of $N$ clusters (or $K$ for the approximate VampPrior), the VMM approaches the VampPrior.
In this sense, the computational complexity of the VMM lies between the VAE and the VampPrior.
For added efficiency, the VMM can turn off gradient computation for eliminated clusters, which is not possible for the VampPrior as it tends to use all $K$ components as \cref{sec:vae-experiments,sec:singe-cell} demonstrate.

\subsection{Related Work}
\label{subsec:related-work}

Our alternating inference procedure is similar to~\citet{lee_meta-gmvae_2020}, who alternate VI and EM steps for a non-Bayesian GMM in the context of meta-learning.
Our work is distinct from theirs in that we use a Bayesian mixture and the VampPrior to parameterize the cluster center distributions.
Also, our intended applications are the latent clustering of images and scRNA-seq data integration.

\citet{zhou_comprehensive_2022} survey and taxonomize various deep clustering approaches, of which VAE-based methods are a subset.
Our focus on VAEs is driven by their success in single-cell genomics and our hypothesis that these methods will benefit from a more flexible prior given the multimodal nature of biological data.

The DLGMM~\citep{nalisnick_approximate_2016} uses a GMM prior~\cref{eq:gmm} but with fixed component means (at equally spaced points on the line) and variances (one), learning only the prior weights $\pi$~\cref{eq:mixing-prior}.
They have $q_{\phi}(z_{ij};x_i)$ for $j\in[K]$, which requires evaluating the decoder $K$ times and averaging the result according to $\pi_i \sim q_{\phi}(\pi_i;x_i)$, a Kumaraswamy stick-breaking variational family~\citep{nalisnick_stick-breaking_2017}.
\citet{stirn_new_2019} show this construction is non-exchangeable and thereby has limited approximation capacity.
We attempted to scale their approach beyond one-dimensional $z_i$\('\)s with limited success but include their reported performance.

The GMVAE~\citep{dilokthanakul_deep_2016} is,
\begin{align*}
    z_i &\sim \N(0,I),
    &c_i \sim \text{Categorical}(\pi),\\
    y_i|z_i,c_i &\sim \N(f_{c_i}(z_i;\theta_{c_i})),
    &x_i|y_i \sim p(x_i;f_x(y_i;\theta_x))
\end{align*}
for observed $x_i$, utilizing $K$ decoders ($f_1,\hdots,f_K$).
Their chosen variational family, $q(z_i,y_i,c_i) \triangleq$
\begin{align*}
    q(z_i;g_z(x_i;\phi_z)) q(y_i;g_y(x_i;\phi_y)) q(c_i|z_i,y_i),
\end{align*}
sets $q(c_i|z_i,y_i)$ to the true posterior.
Given their computational complexity, we rely on their reported performance.

VaDE~\citep{jiang_variational_2017} is closest to our proposals.
They optimize a GMM for the VAE's $p(z)$,
\begin{align*}
    &c_i \sim \text{Categorical}(\pi), &z_i|c_i\sim \N(\mu_{c_i},\Sigma_{c_i}).
\end{align*}
They define $q(c_i;x_i)$ as we do in our E-step and $q_{\phi}(z_i;x_i)$ in the usual way and optimize their ELBO \wrt both variational and prior parameters.
For small $K$ fixed a-priori in the non-Bayesian setting, VaDE cannot automatically discover an appropriate number of clusters (as our Bayesian GMM and VMM do), often utilizing all available, and is highly susceptible to poor initializations.
To avoid poor initializations, they pre-train a deterministic autoencoder, fit a GMM to the latent space, and use those GMM parameters to initialize VaDE's GMM parameters.
Both our GMM and VMM do not require such pre-training.

\begin{table*}[t!]
    \caption{Clustering performance on benchmark datasets. Reported results appear in the top partition, where a `--' denotes missing. We collected all results in the bottom partition. \statTest}
    \label{tab:vae-clustering-performance}
    \centering
    \adjustbox{width=\textwidth}{\begin{tabular}{l|ccc|ccc}
\toprule
& \multicolumn{3}{c|}{mnist w/ $\dim(z) = 10$} & \multicolumn{3}{c}{fashion mnist w/ $\dim(z) = 30$} \\
Method & Utilized clusters & Accuracy & NMI & Utilized clusters & Accuracy & NMI \\
\midrule
DLGMM~\citep{nalisnick_approximate_2016} & 10 & 91.58 & -- & -- & -- & -- \\
GMVAE~\citep{dilokthanakul_deep_2016} & 10 $\pm$ 0.00 & 0.778 $\pm$ 0.06 & -- & -- & -- & -- \\
 & 16 $\pm$ 0.00 & 0.851 $\pm$ 0.02 & -- & -- & -- & -- \\
 & 30 $\pm$ 0.00 & 0.928 $\pm$ 0.02 & -- & -- & -- & -- \\
\midrule
VampPrior~\citep{tomczak_vae_2018} & 100 $\pm$ 0.00 & 0.945 $\pm$ 0.011 & 0.627 $\pm$ 0.003 & 100 $\pm$ 0.00 & \textbf{0.775 $\pm$ 0.016} & 0.529 $\pm$ 0.003 \\
VaDE~\citep{jiang_variational_2017} & 10 $\pm$ 0.00 & 0.857 $\pm$ 0.067 & 0.832 $\pm$ 0.033 & 5.5 $\pm$ 2.51 & 0.352 $\pm$ 0.066 & 0.499 $\pm$ 0.113 \\
Our GMM & 10.7 $\pm$ 2.31 & 0.902 $\pm$ 0.043 & 0.876 $\pm$ 0.013 & 16.1 $\pm$ 3.14 & 0.513 $\pm$ 0.046 & 0.509 $\pm$ 0.037 \\
Our VMM & 13.9 $\pm$ 2.13 & \textbf{0.960 $\pm$ 0.006} & \textbf{0.899 $\pm$ 0.015} & 16.5 $\pm$ 2.92 & 0.712 $\pm$ 0.009 & \textbf{0.653 $\pm$ 0.015} \\
\bottomrule
\end{tabular}
}
\end{table*}

\citet{hrovatin_integrating_2024} and \citet{lopez_destvi_2022} leverage the VampPrior for scRNA-seq and spatial transcriptomic data analysis, respectively.
In contrast, our VMM, a DP-approximating GMM, leverages ideas from the VampPrior to parameterize the variational cluster distributions.

Both the VMM and VampPrior approximate the Empirical Bayes solution (i.e.\ the aggregate posterior), but the VMM provides additional representational flexibility and contains the VampPrior as a special case.
In particular, the precision matrices $\Lambda_j$ in our DP-GMM provide an additional degree of freedom when learning cluster widths.
This flexibility proved critical for the VMM's clustering performance.
The VampPrior burdens the encoder's amortized parameter map with the conflicting tasks of producing both low variance posteriors $q_\phi(z;x)$ when operating on data $x$ to drive predictive performance and high variance prior components $q_\phi(z;u)$ when operating on pseudo-inputs $u$ to drive the marginal likelihood.
To the best of our knowledge, this paradox has not yet been recognized in the literature.
The VMM resolves this paradox by learning the additional covariance terms $\Lambda^{-1}_j$.

\section{VAE EXPERIMENTS}
\label{sec:vae-experiments}

While our intended application is scRNA-seq integration, we first test our methods on the familiar MNIST and Fashion MNIST datasets.
We rescale images to $[-1,1]$, set~\cref{eq:vae-likelihood} to $\N(x_i|f(z_i;\theta),\sigma^2 I)$, and learn $\sigma^2$ during VAE inference.
Following~\citet{jiang_variational_2017}, we use a three-layer MLP with hidden dimensions [500, 500, 2000] for the encoder and the reverse for the decoder, but anecdotally report that our methods had strong clustering performance for both smaller MLPs and small CNNs.
Our variational $q(z;g(x;\phi))$ has full rank covariance.
We set $K=10$ for VaDE (as they do) and $K=100$ for the VampPrior, our GMM, and the VMM\@.
Reported results average over 10 different trials.
Within each trial, data folds and neural network initializations are identical for the VampPrior, VaDE, our GMM, and the VMM\@.
\Cref{sec:vae-experiments-sup} contains additional details.

\subsection{Clustering Performance}
\label{subsec:clustering-performance}

\Cref{tab:vae-clustering-performance} compares clustering performance for VAE-based methods.
Because we are in the unsupervised setting, we measure clustering performance on the combined training and validation sets after early stopping on validation set performance (see \cref{sec:vae-experiments-sup} for details).
Computing accuracy for models whose cluster count differs from the number of ground truth classes necessitates matching cluster assignments to ground truth labels.
To accomplish this, we assign cluster members the label of its most probable member.
However, this method trivially achieves 100\% accuracy when placing an atom on each datapoint.
Thus, the exact VampPrior (the aggregate posterior) over all $N$ data, having a component for each datapoint, will also trivially achieve nearly perfect accuracy.
By contrast, normalized mutual information (NMI) penalizes a model for using too many clusters since it compares a model's cluster assignments directly against labels.
Therefore, NMI is a more appropriate metric for evaluating clustering performance in this setting.
\citet{yang_clustering_2020,winter_common_2022} report NMI values for vanilla GMMs and DP-GMMs that are far lower than our VMM's.
We also calculated the Adjusted Rand Index (ARI) and found it had perfect Spearman correlation with NMI, so we report NMI only.
We report accuracy for familiarity and consistency with other work.

For both datasets, our VMM identifies an appropriate number of clusters (knowing there are 10 labeled classes) and always has the best NMI\@.
For Fashion MNIST, the VampPrior has the best accuracy, but uses all available clusters as it did for MNIST\@.
In contrast, our VMM uses a reasonable number of clusters and has the best accuracy for Fashion MNIST as reported by \href{https://paperswithcode.com/sota/image-clustering-on-fashion-mnist}{papers-with-code}.
Our VaDE implementation's maximum accuracy of 0.936 on MNIST is nearly identical to their reported value of 0.945, suggesting our implementation is correct.
Interestingly, VaDE's clustering performance does not generalize to Fashion MNIST\@.
We suspect their pre-training procedure failed to provide a good initialization because of higher latent dimensions and/or Fashion MNIST's increased heterogeneity.
Our Bayesian GMM uses no pre-training and outperforms VaDE on both datasets, confirming the benefit of our choice to use a DP-approximating mixture.
The VMM outperforming our GMM corroborates the benefit of using VampPrior concepts to fit variational cluster centers in the GMM as opposed to non-amortized point estimates.

\begin{table*}[ht!]
    \caption{VAE model performance for different priors on Fashion MNIST. \statTest}
    \label{tab:vae-model-performance}
    \centering
    \adjustbox{width=0.75\textwidth}{\begin{tabular}{l|lcccccc}
\toprule
Prior & $\log p(x)$ & $\E_q[\log p(x|z)]$ & $\dkl(q(z|x)||p(z))$ & $\dkl(q(z)||p(z))$ & $\mathbb{I}[z;n]$ \\
\midrule
$\mathcal{N}(0,I)$ & 46.9 $\pm$ 2.48 & 59.1 $\pm$ 2.98 & \textbf{28 $\pm$ 0.15} & \textbf{18.8 $\pm$ 0.16} & 9.21 $\pm$ 0.00 \\
VampPrior & \textbf{165 $\pm$ 1.42} & \textbf{197 $\pm$ 1.90} & 50.6 $\pm$ 0.46 & 41.4 $\pm$ 0.46 & 9.21 $\pm$ 0.01 \\
Our GMM & 53.1 $\pm$ 7.88 & 65.2 $\pm$ 8.44 & \textbf{28.1 $\pm$ 1.21} & \textbf{18.9 $\pm$ 1.21} & 9.21 $\pm$ 0.00 \\
Our VMM & 104 $\pm$ 7.50 & 124 $\pm$ 8.90 & 35.2 $\pm$ 1.59 & 25.8 $\pm$ 2.19 & 9.21 $\pm$ 0.01 \\
\bottomrule
\end{tabular}
}
\end{table*}

\begin{figure}[hb!]
    \centering
    \includegraphics[width=\columnwidth]{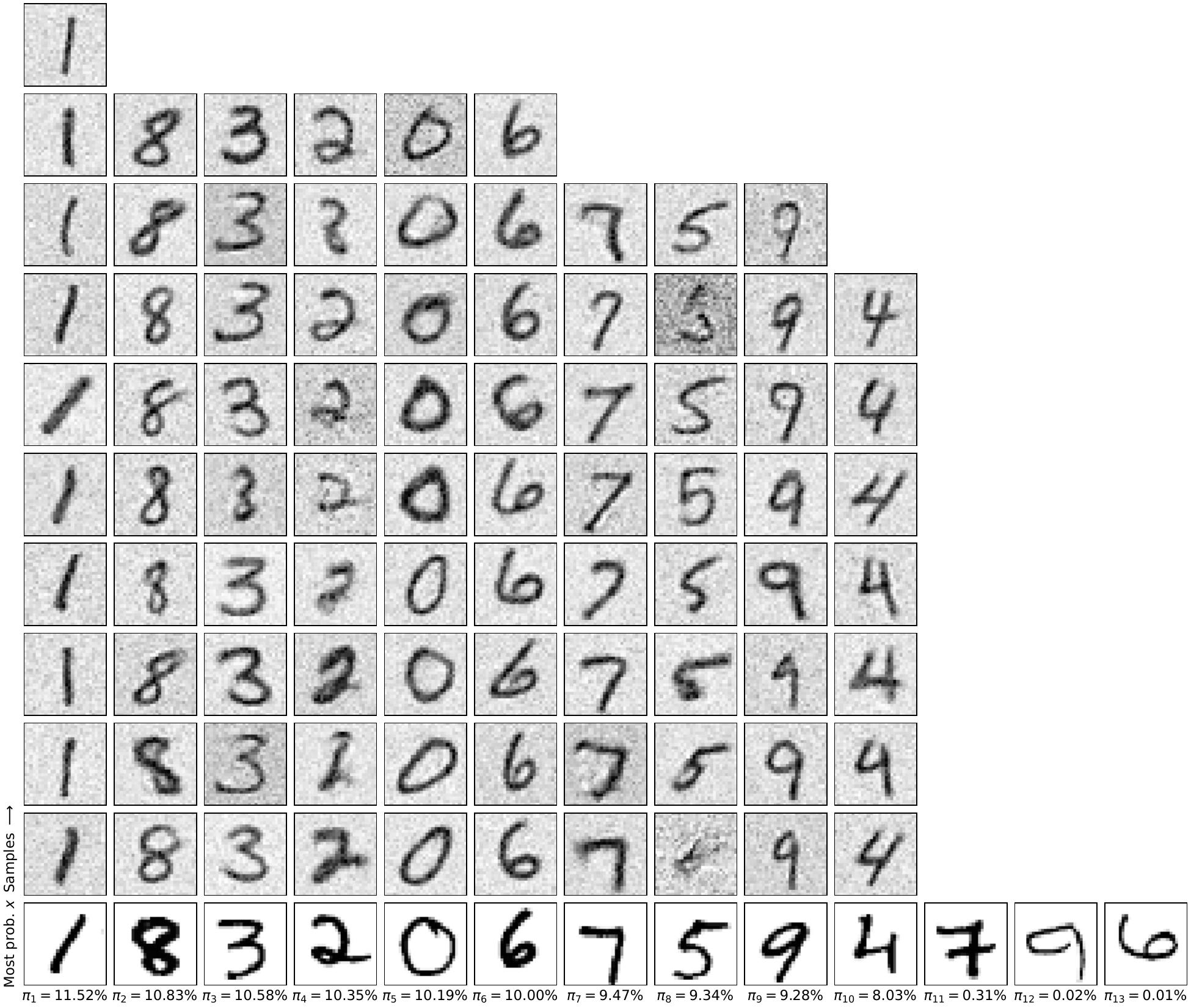}
    \caption{VMM prior predictive samples for MNIST. The number of columns equals the number of utilized clusters. The bottom row shows the data with the highest probability of belonging to the cluster, under which we print the cluster's probability $\pi_j$. The rows above are samples from the corresponding cluster. We sample $\text{round}(10 \cdot \pi_j / \max(\pi))$ images from each component $j$ to help visualize cluster proportions. No samples indicate this value rounded to zero.}
    \label{fig:prior-predictive-samples-main}
\end{figure}

In \cref{fig:prior-predictive-samples-main}, each of the first 10 clusters (columns) represents one of the 10 digits and
comprise 99.6\% of the data according to the VMM's cluster proportions $\pi$ (printed below each column).
The next largest cluster (11th column) has $\pi_j$ = 0.3\% and captures an alternative way to write a `7', known as the slashed 7.
An anthropologist unfamiliar with our numerical representation system would likely find the 11th cluster interesting.
Utilizing 11 of the 100 available clusters to represent 99.9\% of MNIST suggests the VMM excels at identifying a semantically representative set of clusters.
We see similar results for Fashion MNIST in \cref{sec:vae-experiments-sup}.

Comparing samples from the marginal of a VMM and a VampPrior (\cref{fig:prior-predictive-samples-main} and supplement \cref{fig:prior-predictive-samples-sup}), we see that the diversity of samples from the VMM's cluster components is markedly higher than those from the VampPrior.
The VampPrior uses $q_{\phi}(z;u_j)$ to fully specify a cluster, whereas the VMM uses $q_{\phi}(\mu;u_j)$ to specify just the cluster's mean and uses $\Lambda_j$ to specify its width in each latent dimension, affording it the ability to generate more diversity from a single cluster and thereby use fewer clusters.
While diverse, each cluster in the VMM produces semantically similar samples.

\subsection{Model Comparison}
\label{subsec:model-comparison}

In \cref{tab:vae-model-performance}, we report the marginal likelihood, the negative distortion and rate~\citep{alemi_fixing_2018}, and the marginal KL divergence and mutual information (MI) between latent embedding $z$ and index code $n$~\citep{hoffman_elbo_2016} as measured on the validation fold.
These quantities obey,
\begin{align}
    \nonumber \underbrace{\log p(x)}_\text{log marginal} &\geq \underbrace{\E_q[\log p(x|z)]}_{-\text{distortion}} - \underbrace{\dkl(q(z|x)||p(z))}_\text{rate} \\
    \label{eq:marginal-kl} \underbrace{\dkl(q(z|x)||p(z))}_\text{rate} &= \underbrace{\dkl(q(z)||p(z))}_\text{marginal KL} + \underbrace{\mathbb{I}[z;n]}_\text{MI}.
\end{align}
Unsurprisingly, the VampPrior has the highest evidence since it is, by design, approximately optimal \wrt the ELBO (\cref{subsec:vampprior}).
This approximation gap narrows as $K\rightarrow N$. \citet{tomczak_vae_2018} observe ELBO improvements as $K$ increases.
In contrast, our VMM works to reduce the VampPrior's component usage to achieve a more interpretable clustering and therefore should have a lower marginal likelihood.

\citet{hoffman_elbo_2016} show that $\mathbb{I}[z;n] \leq \log N$ and approaches this upper bound (9.21 for validation set size) for $p \geq 10$.
Thus, by \cref{eq:marginal-kl}, the model with the highest rate will also have the largest marginal KL divergence.
The VampPrior's ELBO maximization trades an increased rate for superior distortion, leading to a rather paradoxical result:
despite the VampPrior setting $p(z)$ to approximate the aggregate posterior $q(z)$, it has the largest $\dkl(q(z)||p(z))$, suggesting it is ineffective at regulating the structure of the latent embedding.
\Cref{tab:vae-model-performance} shows the VMM has substantially more modelling power than the GMM, despite utilizing the same number of clusters (\cref{tab:vae-clustering-performance}) and exhibits performance commensurate with it lying between the extremes of the unimodal standard normal and the VampPrior.

\section{SINGLE CELL EXPERIMENTS}
\label{sec:singe-cell}

\begin{table*}[t]
    \caption{scRNA-seq clustering and integration performance. \statTest}
    \label{tab:sc-performance}
    \adjustbox{width=\textwidth}{\begin{tabular}{lll|ccc|ccc}
\toprule
 & & & \multicolumn{3}{c|}{Clustering Performance} & \multicolumn{3}{c}{\citet{luecken_benchmarking_2022} Metrics} \\
Dataset & Model & Prior & Utilized Clusters & Accuracy & NMI & Batch correction & Bio conservation & Total\\
\midrule
\multirow[t]{5}{*}{cortex} & \href{https://docs.scvi-tools.org/en/stable/user_guide/models/scvi.html}{scVI tools} & $\mathcal{N}(0,I)$ & -- & -- & -- & -- & 0.696 $\pm$ 0.020 & -- \\
 & \multirow[t]{4}{*}{scVI (our code)} & $\mathcal{N}(0,I)$ & -- & -- & -- & -- & 0.694 $\pm$ 0.019 & -- \\
 &  & VampPrior & 100 $\pm$ 0.00 & \textbf{0.880 $\pm$ 0.022} & 0.492 $\pm$ 0.005 & -- & \textbf{0.760 $\pm$ 0.007} & --  \\
 &  & GMM & 19.7 $\pm$ 1.64 & 0.851 $\pm$ 0.037 & \textbf{0.652 $\pm$ 0.016} & -- & 0.740 $\pm$ 0.007 & -- \\
 &  & VMM & 28.9 $\pm$ 2.88 & 0.833 $\pm$ 0.041 & \textbf{0.663 $\pm$ 0.009} & -- & \textbf{0.755 $\pm$ 0.007} & -- \\
 & Harmony & N/A & -- & -- & -- & -- & 0.645 $\pm$ 0.000 & -- \\
 & Scanorama & N/A & -- & -- & -- & -- & 0.645 $\pm$ 0.000 & -- \\
\midrule
\multirow[t]{5}{*}{pbmc} & \href{https://docs.scvi-tools.org/en/stable/user_guide/models/scvi.html}{scVI tools} & $\mathcal{N}(0,I)$ & -- & -- & -- & 0.860 $\pm$ 0.011 & 0.661 $\pm$ 0.020 & 0.741 $\pm$ 0.012 \\
 & \multirow[t]{4}{*}{scVI (our code)} & $\mathcal{N}(0,I)$ & -- & -- & -- & 0.840 $\pm$ 0.013 & 0.669 $\pm$ 0.016 & 0.737 $\pm$ 0.010 \\
 &  & VampPrior & 100 $\pm$ 0.00 & \textbf{0.943 $\pm$ 0.020} & 0.507 $\pm$ 0.003 & \textbf{0.883 $\pm$ 0.007} & 0.740 $\pm$ 0.014 & \textbf{0.797 $\pm$ 0.008} \\
 &  & GMM & 12.2 $\pm$ 1.69 & 0.729 $\pm$ 0.109 & 0.756 $\pm$ 0.020 & 0.866 $\pm$ 0.007 & 0.722 $\pm$ 0.016 & 0.779 $\pm$ 0.009 \\
 &  & VMM & 14.1 $\pm$ 0.74 & \textbf{0.933 $\pm$ 0.014} & \textbf{0.817 $\pm$ 0.030} & \textbf{0.886 $\pm$ 0.009} & 0.739 $\pm$ 0.011 & \textbf{0.798 $\pm$ 0.007} \\
 & Harmony & N/A & -- & -- & -- & 0.868 $\pm$ 0.001 & 0.734 $\pm$ 0.000 & 0.787 $\pm$ 0.001 \\
 & Scanorama & N/A & -- & -- & -- & 0.486 $\pm$ 0.007 & \textbf{0.756 $\pm$ 0.009} & 0.648 $\pm$ 0.005 \\
\midrule
\multirow[t]{5}{*}{split-seq} & \href{https://docs.scvi-tools.org/en/stable/user_guide/models/scvi.html}{scVI tools} & $\mathcal{N}(0,I)$ & -- & -- & -- & 0.864 $\pm$ 0.005 & 0.598 $\pm$ 0.008 & 0.705 $\pm$ 0.005 \\
 & \multirow[t]{4}{*}{scVI (our code)} & $\mathcal{N}(0,I)$ & -- & -- & -- & 0.875 $\pm$ 0.006 & 0.612 $\pm$ 0.007 & 0.717 $\pm$ 0.004  \\
 &  & VampPrior & 100 $\pm$ 0.00 & \textbf{0.865 $\pm$ 0.013} & 0.536 $\pm$ 0.004 & \textbf{0.876 $\pm$ 0.005} & 0.638 $\pm$ 0.010 & \textbf{0.733 $\pm$ 0.006} \\
 &  & GMM & 14.1 $\pm$ 1.85 & 0.636 $\pm$ 0.067 & 0.600 $\pm$ 0.024 & \textbf{0.876 $\pm$ 0.008} & 0.631 $\pm$ 0.006 & 0.729 $\pm$ 0.006 \\
 &  & VMM & 28.4 $\pm$ 2.72 & \textbf{0.827 $\pm$ 0.060} & \textbf{0.641 $\pm$ 0.008}& \textbf{0.878 $\pm$ 0.004} & 0.639 $\pm$ 0.004 & \textbf{0.734 $\pm$ 0.003} \\
 & Harmony & N/A & -- & -- & -- & 0.790 $\pm$ 0.002 & \textbf{0.650 $\pm$ 0.002} & 0.706 $\pm$ 0.002 \\
 & Scanorama & N/A & -- & -- & -- & 0.552 $\pm$ 0.000 & 0.632 $\pm$ 0.000 & 0.600 $\pm$ 0.000 \\
\midrule
\multirow[t]{5}{*}{lung atlas} & \href{https://docs.scvi-tools.org/en/stable/user_guide/models/scvi.html}{scVI tools} & $\mathcal{N}(0,I)$ & -- & -- & -- & \textbf{0.620 $\pm$ 0.006} & 0.608 $\pm$ 0.015 & 0.613 $\pm$ 0.010 \\
 & \multirow[t]{4}{*}{scVI (our code)} & $\mathcal{N}(0,I)$ & -- & -- & -- & 0.616 $\pm$ 0.006 & 0.600 $\pm$ 0.008 & 0.607 $\pm$ 0.006 \\
 &  & VampPrior & 100 $\pm$ 0.00 & \textbf{0.627 $\pm$ 0.022} & 0.548 $\pm$ 0.010 & 0.538 $\pm$ 0.014 & \textbf{0.688 $\pm$ 0.015} & \textbf{0.628 $\pm$ 0.012} \\
 &  & GMM & 10.2 $\pm$ 1.32 & 0.305 $\pm$ 0.089 & 0.447 $\pm$ 0.037 & 0.611 $\pm$ 0.007 & 0.628 $\pm$ 0.019 & \textbf{0.621 $\pm$ 0.013} \\
 &  & VMM & 20 $\pm$ 6.32 & 0.580 $\pm$ 0.040 & \textbf{0.641 $\pm$ 0.020} & 0.562 $\pm$ 0.012 & 0.674 $\pm$ 0.012 & \textbf{0.629 $\pm$ 0.008} \\
 & Harmony & N/A & -- & -- & -- & 0.553 $\pm$ 0.003 & 0.678 $\pm$ 0.002 & \textbf{0.628 $\pm$ 0.002} \\
 & Scanorama & N/A & -- & -- & -- & 0.328 $\pm$ 0.006 & \textbf{0.694 $\pm$ 0.009} & 0.548 $\pm$ 0.007 \\
\bottomrule
\end{tabular}
}
\end{table*}

We now integrate the VampPrior, our GMM, and our VMM into scVI~\citep{lopez_deep_2018}.
In its default configuration, scVI models scRNA-seq read counts using amortized VI and a DLVM with generative process,
\begin{align*}
    z_i \sim \N(0,I), \quad x_i|z_i,l_i,s_i \sim p\big(x_i;f_{\theta}(z_i,l_i,s_i)\big).
\end{align*}
For each cell $i\in[N]$, we observe the read counts for $G$ genes $x_i \in \mathbb{N}^G$, library size $l_i=\sum_j x_{ij}$ as the total counts for cell $i$, and an integer batch (e.g.~donor or patient) identifiers $s_i$.
Here, $f_{\theta}(z_i,l_i,s_i)$ is a neural network that parameterizes a zero-inflated negative binomial (ZINB) distribution for all datasets except the lung atlas dataset, for which the scVI authors \href{https://docs.scvi-tools.org/en/stable/tutorials/notebooks/scrna/harmonization.html}{recommend a negative binomial}.
scVI defines the variational family $q_{\phi}(z_i;x_i,s_i)$ as a normal distribution with diagonal covariance.
Thus, scVI is simply a VAE with a $\N(0,I)$ prior and a ZINB likelihood except that it additionally uses $s_i$ to parameterize the variational and likelihood distributions.

Single-cell RNA-seq integration seeks a latent representation for biological variation that is disentangled from batch-specific technical variation.
Batch variables $s$ serve only as batch-specific calibration factors for the data generating distribution.
Giving both the encoder and decoder access to $s_i$ encourages scVI to reserve $z_i$\('\)s limited channel capacity for just biological variability, thus promoting disentanglement.
Ideally, the encoder uses $s_i$ to map $x_i$ onto $z_i$, a batch-effect-free subspace, from which the decoder can reconstruct $x_i$ given its access to $s_i$.

Incorporating our Bayesian GMM and VMM into scVI follows \cref{sec:methods}, with one exception for the VampPrior and our VMM\@.
Encoder network inputs $x_i$ and $s_i$ are non-negative integers and one-hot encoded batch IDs, respectively.
For the part of each pseudo-input $u$ corresponding to counts $x$, we relax integer constraints and fit a non-negative real $G$-vector via softplus-reparameterization.
For the part of each pseudo-input corresponding to batch $s$, we fit a probability vector via a softmax-reparameterization.

We use \href{https://docs.scvi-tools.org/en/stable/api/datasets.html}{scVI's API} to download the cortex, PBMC, and lung atlas datasets and process them according to \citet{lopez_deep_2018}.
For each dataset, we use scVI's recommended architecture, latent dimension, and learning rate for amortized VI ($\gamma_1$ in \cref{alg:aternating-inference}).
Reported results are averages from 10 different trials.
Within each trial, data folds are identical for all methods and neural network initializations are identical for our scVI implementations.
See \cref{sec:single-cell-sup} for additional details.

The left set of metrics in \cref{tab:sc-performance} compares clustering performance.
On all datasets the VMM achieves the best clustering performance according to NMI, suggesting strong agreement with the datasets' cell type annotations.
The VampPrior's tendency to use all available clusters allows it to achieve good accuracy but hamstrings its NMI, suggesting over-clustering.
Since the $\N(0,I)$ doesn't explicitly cluster, we do not report its clustering performance.

\Cref{tab:sc-performance} (right) compares scRNA-seq integration using \citet{luecken_benchmarking_2022}\('\)s benchmarking suite, which produces three summary scores: ``batch correction'' assesses the removal of technical variation, ``bio conservation'' assesses the preservation of biological variation, and ``total'' is a weighted combination thereof.
Batch correction and bio conservation scores are each a composite of five different metrics (supplement \cref{tab:sc-integration-performance}).
\citet{luecken_benchmarking_2022} scale all metrics to $[0,1]$ such that larger values denote better performance.

\begin{figure*}[p]
    \centering
    \centerline{\includegraphics[width=\textwidth]{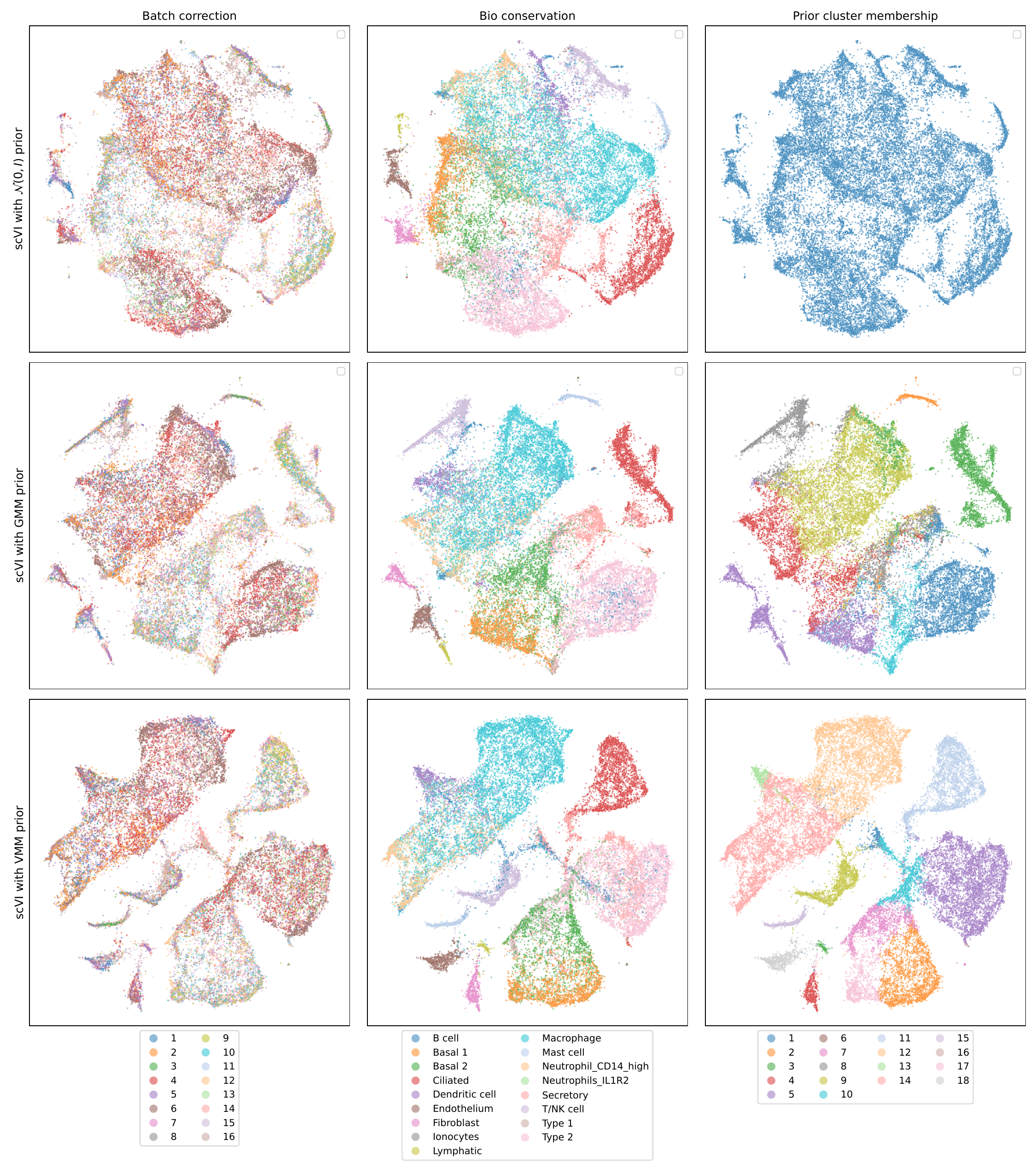}}
    \caption{\mde{lung atlas}}
    \label{fig:mde-lung-atlas}
\end{figure*}

The cortex dataset only has a single batch and thus lacks batch correction scores.
Because of this, Harmony and Scanorama scores are from using PCA directly and thus identical.
Our VMM has (or is tied with) the best total score on all datasets, suggesting the VMM is near state of the art.
The VMM also has or matches the top batch correction score on each dataset except the lung atlas dataset, which has 16 donors and 17 cell types--a true atlas-level integration task.
Examining the batch correction component scores (supplement \cref{tab:sc-integration-performance}), we find that PCR (principal component regression) is the batch correction metric dragging down the VMM's batch correction score.
PCR measures the relative reduction in total variance explained by the batch variable $S$ for the integrated representation $Z$ \wrt the raw count data $X$ as $\frac{\text{Var}(X|S) - \text{Var}(Z|S)}{\text{Var}(X|S)}$.
The score attains its maximal value when $\text{Var}(Z|S)=0$, which is an unreasonable expectation for the lung atlas dataset since each donor has a different distribution of cell types with some donors only having a subset of the possible cell types.
Therefore, we expect batch variable $S$ to explain some amount of representation $Z$.
In fact, it is concerning that scVI with a $\N(0,I)$ prior does so well here.
This possibly suggests \emph{over integration} as evidenced by its poorly structured Minimum-Distortion Embedding (MDE) plot in \cref{fig:mde-lung-atlas}.
In contrast, the VMM has a well-structured MDE and finds meaningful clusters when comparing the cell-type annotations to VMM cluster assignments, supporting our case for simultaneous integration and clustering.
The GMM's PCR metric and its qualitative MDE structure lies between those for the $\N(0,I)$ and VMM\@.
\Cref{sec:single-cell-sup} contains the remaining MDE comparison plots.

To confirm the VMM's batch performance scores on lung atlas were an artifact of the dataset, we use the SPLiT-seq dataset~\citep{shabestari_absence_2022} as a control, because there every cell is sequenced twice using two slightly different sequencing library preparation steps.
Thus, the cells' biological variability should be identical across the two batches (one for each library preparation).
Indeed, we find the VMM is the top performer on this dataset, allaying concerns it hampered batch correction on the lung atlas data.

\section{CONCLUSIONS}
\label{sec:conclusions}

Advocating for the simultaneous integration and clustering of scRNA-seq, we developed the VampPrior Mixture Model (VMM) and an inference procedure that are adaptable to any DLVM model with continuous latent variables.
The VMM produces well-clustered latent representations for both scRNA-seq and natural images.

The VMM outperforms all VAE-based clustering methods and attains highly competitive image clustering performance relative to all methods.
To the best of our knowledge, we are the first to observe the VampPrior's paradoxical marginal KL divergence behavior and its ineffectiveness at finding semantically meaningful latent structure.

Replacing scVI's~\citep{lopez_deep_2018} $\N(0,I)$ prior with the VMM significantly elevates its scRNA-seq integration performance, improving both batch correction and biological conservation.
There are various extensions of scVI for other biological data modalities, which use $\N(0,I)$ priors in their generative processes.
We conjecture these methods would derive substantial benefit by replacing the $\N(0,I)$ with the VMM and employing our alternating inference procedure.
Furthermore, these methods would inherit the VMM's clustering ability, thereby simplifying downstream analysis.

The VMM is adept at clustering data into a semantically meaningful number of clusters when tuned to match the number of ground truth classes.
We hope that the ability to control the VMM's cluster granularity via the batch size and/or the Empirical Bayes learning rate will help data scientists make novel discoveries in settings where ground truth labels are unknown.

\subsubsection*{Acknowledgements}
We thank David M. Blei for his helpful comments and the anonymous reviewers for their insightful questions and suggestions.
This work was made possible by support from the MacMillan Family via the MacMillan Center for the Study of the Non-Coding Cancer Genome at the
New York Genome Center. 
This material is based upon work supported by the National Science Foundation under CAREER Grant No. DBI2146398.
Any opinions, findings, and conclusions or recommendations expressed in this material are those of the authors and do not necessarily reflect the views of the National Science Foundation.

\printbibliography

\newpage
\appendix
\onecolumn
\aistatstitle{The VampPrior Mixture Model: Supplementary Materials}
\flushleft

\section{SOFTWARE AVAILABILITY AND COMPUTE INFRASTRUCTURE}
\label{sec:software-availability}
Our code is available at \url{https://github.com/astirn/VampPrior-Mixture-Model}.
We performed all experiments on a laptop with 16GB of RAM connected to an external Nvidia 3090 GPU\@.

\section{VMM OPTIMIZATION}
\label{sec:vmm-sup}

Adapting \cref{alg:aternating-inference} to the VMM results in the following procedure:
\begin{align*}
    \text{sample } & x_1,\hdots,x_M \sim \mathcal{D} \\
    \maximize_{\theta,\phi} & \sum_{i=1}^M \E_{q_{\phi}(z;x_i)}\big[\log p_{\theta}(x_i|z) \big] - \text{KL}\Bigg(q_{\phi}(z;x_i) \Big|\Big| \sum_{j=1}^K \pi_j \E_{q_{\phi}(\mu_j;u_j)}\big[\mathcal{N}(z|\mu_j,\Lambda^{-1}_j)\big] \Bigg) \\
    \text{sample } & z_i \sim q_{\phi}(z;x_i) \ \forall i \in [M] \\
    \text{set } & q(c_i;z_i) \coloneqq \text{softmax}\Big(\log \pi + \E_{q_{\phi}(\mu;u)}\big[\log\mathcal{N}(z_i|\mu,\Lambda)\big]\Big) \ \forall i \in [M] \\
    \maximize_{\alpha,\pi,u,\Lambda} & \E_{q(c;z)q_{\phi}(\mu;u)}\Big[
        \log p(\alpha) + \log p(\pi|\alpha) + \log p(\mu) + \log p(\Lambda) + \log \mathcal{N}(z_i|\mu,\lambda)
    \Big],
\end{align*}
where $p(\alpha)$, $p(\pi|\alpha)$, $p(\mu)$, and $p(\Lambda)$ correspond to \cref{eq:concentration-prior,eq:mixing-prior,eq:centers-prior-gmm,eq:precisions-prior}.

\section{VAE EXPERIMENTS}
\label{sec:vae-experiments-sup}

\begin{figure}[hb!]
    \centering
    \includegraphics[width=0.7\textwidth]{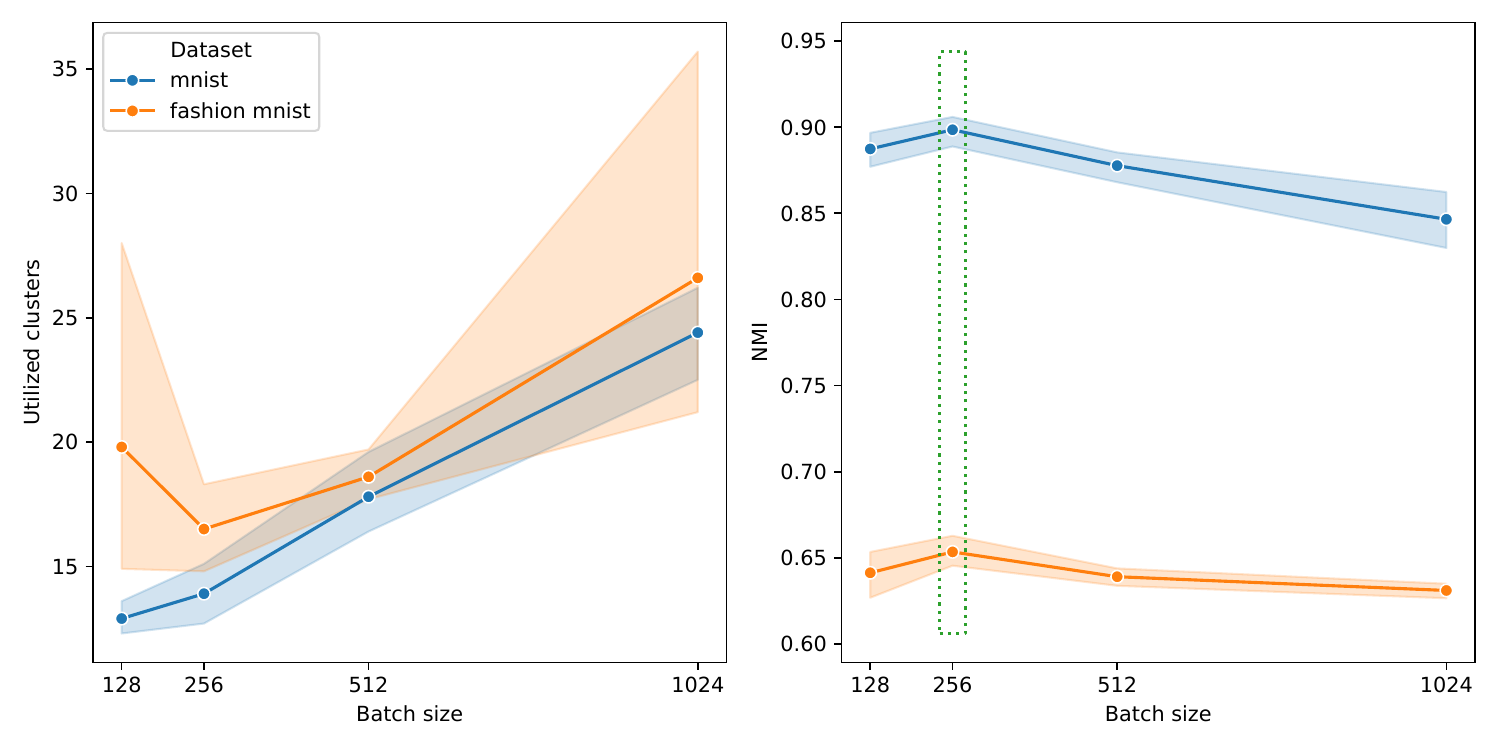}
    \caption{VMM cluster utilization and NMI performance for different batch sizes. Shading denotes 95\% CIs. The dotted green rectangle marks the peak NMI performance between class labels and cluster assignments.}
    \label{fig:clustering-tuning}
\end{figure}

\Cref{subsec:aternating-inference} discusses how the VMM's batch size setting serves as a simple proxy for adjusting the DP's concentration, avoiding the need to adjust hyper-prior parameters.
Increasing the data available to the Empirical Bayes EM step in \cref{alg:aternating-inference} weakens the effect of the hyper-priors resulting in more clusters being utilized.
We employ this strategy for the MNIST and Fashion MNIST experiments in \cref{sec:vae-experiments}.
For various batch size settings, we compute how many clusters the VMM utilizes $|\{\forall~i\in[N]: \text{arg max}_j~q(c_i=j|x_i)\}|$ and the NMI between class labels and cluster assignments $\text{arg max}_j~q(c_i=j|x_i)$.
Since MNIST and Fashion MNIST have the same number of true classes and approximately uniform class distributions, it is reassuring that batch size's effect on cluster utilization is similar (\cref{fig:clustering-tuning}, left) and that NMI peaks in both cases with a batch size of 256 (\cref{fig:clustering-tuning}, right), which we use for all VAE experiments in \cref{sec:vae-experiments}.

We use Adam~\citep{kingma_adam_2014} with a 1e-4 learning rate for both VI ($\gamma_1$, \cref{alg:aternating-inference}) and Empirical Bayes ($\gamma_2$, \cref{alg:aternating-inference}) steps.
Stochastic gradient ascent's appearance in \cref{alg:aternating-inference} is for illustrative purposes only.
Following \citet{jiang_variational_2017}, we use ReLU activations for all hidden layers.
For \cref{tab:vae-clustering-performance}, we allow all models to train for a maximum of 10,000 epochs, which no model comes close to using since we stop training early if the validation set's NMI has not improved for 100 epochs.
For \cref{tab:vae-model-performance}, we allow all models to train for a maximum of 500 epochs, but stop training early if the validation set's ELBO has not improved for 20 epochs.
In both cases, we restore the model's parameters from the epoch with the best validation set metric of interest.
Please see our provided code for additional implementation details.

\begin{figure}
    \centering
    \begin{subfigure}[b]{\textwidth}
        \centering
        \includegraphics[width=\textwidth]{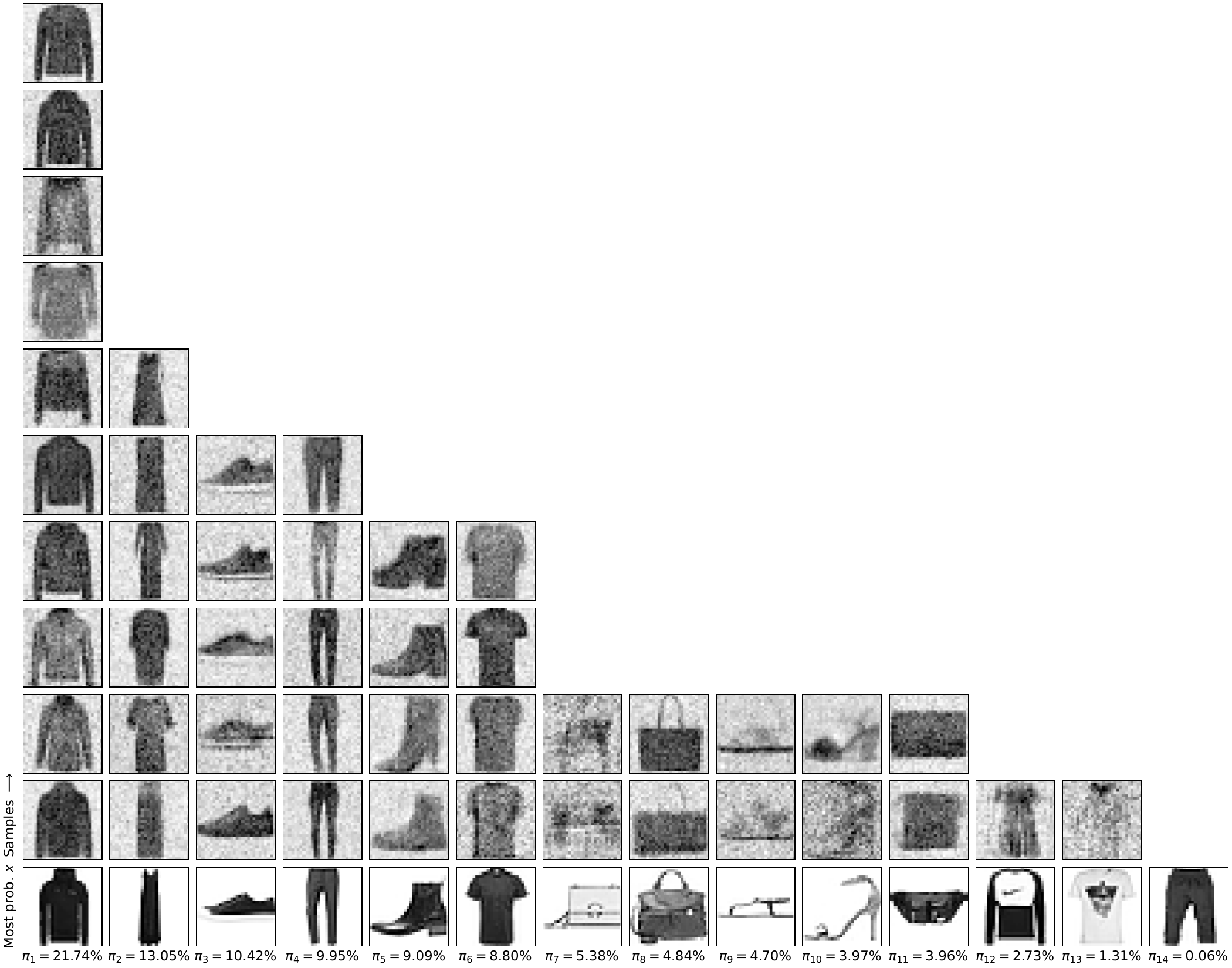}
        \caption{VMM prior predictive samples for Fashion MNIST}
    \end{subfigure}
    \\
    \begin{subfigure}[b]{\textwidth}
        \centering
        \includegraphics[width=\textwidth]{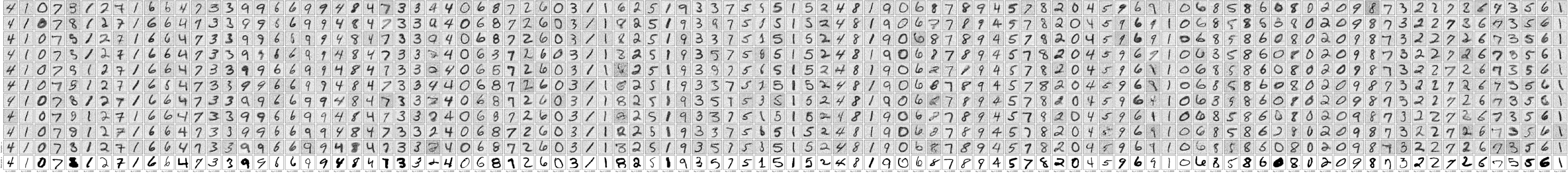}
        \caption{VampPrior predictive samples for MNIST}
    \end{subfigure}
    \\
    \begin{subfigure}[b]{\textwidth}
        \centering
        \includegraphics[width=\textwidth]{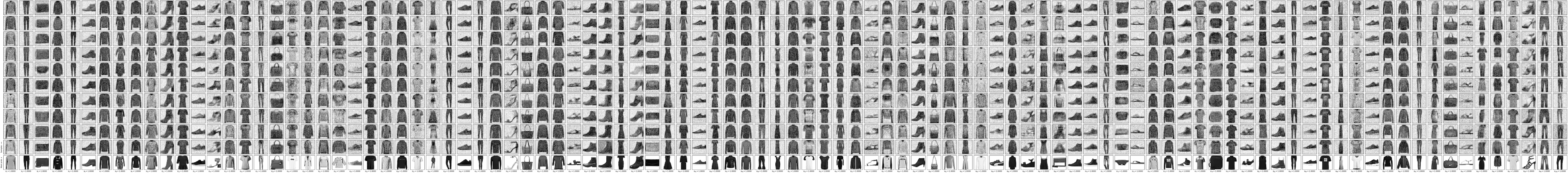}
        \caption{VampPrior predictive samples for Fashion MNIST}
    \end{subfigure}
    \caption{Prior predictive samples. The number of columns equals the number of utilized clusters. The bottom row shows the data with the highest probability of belonging to the cluster, under which we print the cluster's probability $\pi_j$. The rows above are samples from the corresponding cluster. We sample $\text{round}(10 \cdot \pi_j / \max(\pi))$ images from each component $j$ to help visualize cluster proportions. No samples indicate this value rounded to zero. The VampPrior has uniform prior class probabilities since it specifies $\pi_j=K^{-1}$ and does not fit $\pi$ during inference.}
    \label{fig:prior-predictive-samples-sup}
\end{figure}

\Cref{fig:prior-predictive-samples-sup} complements \cref{fig:prior-predictive-samples-main} from the main text.
Panel (a), shows the VMM's prior predictive samples for the Fashion MNIST dataset.
Here, the VMM uses 13 clusters to represent 99.9\% of the data, which has 10 ground truth classes.
The VMM's most populous cluster (column 1) seemingly combines the `pullover' and `coat' classes into an outerwear cluster.
Additionally, the VMM splits the `bag' class into a handbags cluster (column 7) and shoulder bags (column 8) and the `sandal' class into flat (column 9) and high-heeled (column 10) variants.
The VMM also seems to partition graphic variants of the `pull-over' and `t-shirt/top' classes into a separate clusters (columns 12 and 13).
Not only is this clustering semantically justifiable, it is also best in class for Fashion MNIST~\citep{yang_clustering_2020}.

\section{SINGLE CELL EXPERIMENTS}
\label{sec:single-cell-sup}

\begin{table}
    \caption{scRNA-seq integration performance. \statTest}
    \label{tab:sc-integration-performance}
    \adjustbox{width=\textwidth}{\begin{tabular}{lll|ccc|ccccc|ccccc}
\toprule
 &  & Metric Type & \multicolumn{3}{c|}{Aggregate score} & \multicolumn{5}{c|}{Batch correction} & \multicolumn{5}{c}{Bio conservation} \\
 &  &  & Batch correction & Bio conservation & Total & Graph connectivity & KBET & PCR comparison & Silhouette batch & iLISI & Isolated labels & KMeans ARI & KMeans NMI & Silhouette label & cLISI \\
Dataset & Model & Prior &  &  &  &  &  &  &  &  &  &  &  &  &  \\
\midrule
\multirow[t]{7}{*}{cortex} & \href{https://docs.scvi-tools.org/en/stable/user_guide/models/scvi.html}{scVI tools} & $\mathcal{N}(0,I)$ & -- & 0.696 $\pm$ 0.020 & -- & -- & -- & -- & -- & -- & 0.599 $\pm$ 0.002 & 0.641 $\pm$ 0.059 & 0.649 $\pm$ 0.040 & 0.595 $\pm$ 0.002 & 0.994 $\pm$ 0.001 \\
 & \multirow[t]{4}{*}{scVI (our code)} & $\mathcal{N}(0,I)$ & -- & 0.694 $\pm$ 0.019 & -- & -- & -- & -- & -- & -- & 0.604 $\pm$ 0.004 & 0.626 $\pm$ 0.050 & 0.650 $\pm$ 0.040 & 0.595 $\pm$ 0.004 & 0.994 $\pm$ 0.001 \\
 &  & VampPrior & -- & \textbf{0.760 $\pm$ 0.007} & -- & -- & -- & -- & -- & -- & \textbf{0.661 $\pm$ 0.005} & \textbf{0.744 $\pm$ 0.028} & \textbf{0.731 $\pm$ 0.011} & \textbf{0.665 $\pm$ 0.004} & \textbf{0.999 $\pm$ 0.000} \\
 &  & GMM & -- & 0.740 $\pm$ 0.007 & -- & -- & -- & -- & -- & -- & 0.638 $\pm$ 0.004 & 0.715 $\pm$ 0.021 & 0.709 $\pm$ 0.010 & 0.641 $\pm$ 0.004 & 0.998 $\pm$ 0.000 \\
 &  & VMM & -- & \textbf{0.755 $\pm$ 0.007} & -- & -- & -- & -- & -- & -- & 0.647 $\pm$ 0.004 & \textbf{0.752 $\pm$ 0.023} & \textbf{0.725 $\pm$ 0.013} & 0.652 $\pm$ 0.004 & 0.999 $\pm$ 0.000 \\
 & Harmony & N/A & -- & 0.645 $\pm$ 0.000 & -- & -- & -- & -- & -- & -- & 0.593 $\pm$ 0.000 & 0.480 $\pm$ 0.000 & 0.552 $\pm$ 0.000 & 0.605 $\pm$ 0.000 & 0.996 $\pm$ 0.000 \\
 & Scanorama & N/A & -- & 0.645 $\pm$ 0.000 & -- & -- & -- & -- & -- & -- & 0.593 $\pm$ 0.000 & 0.480 $\pm$ 0.000 & 0.552 $\pm$ 0.000 & 0.605 $\pm$ 0.000 & 0.996 $\pm$ 0.000 \\
\midrule
\multirow[t]{7}{*}{pbmc} & \href{https://docs.scvi-tools.org/en/stable/user_guide/models/scvi.html}{scVI tools} & $\mathcal{N}(0,I)$ & 0.860 $\pm$ 0.011 & 0.661 $\pm$ 0.020 & 0.741 $\pm$ 0.012 & \textbf{0.893 $\pm$ 0.011} & 0.885 $\pm$ 0.057 & 0.810 $\pm$ 0.042 & \textbf{0.967 $\pm$ 0.003} & 0.743 $\pm$ 0.006 & 0.572 $\pm$ 0.003 & 0.491 $\pm$ 0.062 & 0.659 $\pm$ 0.034 & 0.585 $\pm$ 0.005 & 0.999 $\pm$ 0.000 \\
 & \multirow[t]{4}{*}{scVI (our code)} & $\mathcal{N}(0,I)$ & 0.840 $\pm$ 0.013 & 0.669 $\pm$ 0.016 & 0.737 $\pm$ 0.010 & \textbf{0.895 $\pm$ 0.013} & 0.855 $\pm$ 0.049 & 0.778 $\pm$ 0.054 & 0.959 $\pm$ 0.003 & 0.714 $\pm$ 0.007 & 0.575 $\pm$ 0.003 & 0.514 $\pm$ 0.051 & 0.670 $\pm$ 0.027 & 0.588 $\pm$ 0.003 & 0.999 $\pm$ 0.000 \\
 &  & VampPrior & \textbf{0.883 $\pm$ 0.007} & 0.740 $\pm$ 0.014 & \textbf{0.797 $\pm$ 0.008} & 0.871 $\pm$ 0.024 & 0.916 $\pm$ 0.035 & 0.900 $\pm$ 0.024 & 0.963 $\pm$ 0.003 & \textbf{0.766 $\pm$ 0.007} & \textbf{0.632 $\pm$ 0.006} & 0.629 $\pm$ 0.051 & 0.753 $\pm$ 0.018 & \textbf{0.686 $\pm$ 0.008} & \textbf{1.000 $\pm$ 0.000} \\
 &  & GMM & 0.866 $\pm$ 0.007 & 0.722 $\pm$ 0.016 & 0.779 $\pm$ 0.009 & 0.879 $\pm$ 0.015 & 0.919 $\pm$ 0.012 & 0.829 $\pm$ 0.026 & 0.959 $\pm$ 0.003 & 0.746 $\pm$ 0.005 & 0.611 $\pm$ 0.004 & 0.610 $\pm$ 0.057 & 0.728 $\pm$ 0.026 & 0.659 $\pm$ 0.004 & \textbf{1.000 $\pm$ 0.000} \\
 &  & VMM & \textbf{0.886 $\pm$ 0.009} & 0.739 $\pm$ 0.011 & \textbf{0.798 $\pm$ 0.007} & 0.866 $\pm$ 0.011 & 0.909 $\pm$ 0.032 & \textbf{0.937 $\pm$ 0.021} & 0.961 $\pm$ 0.003 & 0.757 $\pm$ 0.006 & 0.614 $\pm$ 0.003 & 0.657 $\pm$ 0.034 & 0.754 $\pm$ 0.023 & 0.668 $\pm$ 0.007 & \textbf{1.000 $\pm$ 0.000} \\
 & Harmony & N/A & 0.868 $\pm$ 0.001 & 0.734 $\pm$ 0.000 & 0.787 $\pm$ 0.001 & 0.738 $\pm$ 0.001 & \textbf{0.965 $\pm$ 0.004} & 0.908 $\pm$ 0.003 & \textbf{0.967 $\pm$ 0.000} & 0.761 $\pm$ 0.000 & 0.568 $\pm$ 0.000 & 0.764 $\pm$ 0.000 & 0.728 $\pm$ 0.000 & 0.609 $\pm$ 0.000 & 0.999 $\pm$ 0.000 \\
 & Scanorama & N/A & 0.486 $\pm$ 0.007 & \textbf{0.756 $\pm$ 0.009} & 0.648 $\pm$ 0.005 & 0.373 $\pm$ 0.009 & 0.474 $\pm$ 0.029 & 0.695 $\pm$ 0.011 & 0.843 $\pm$ 0.001 & 0.043 $\pm$ 0.000 & 0.585 $\pm$ 0.000 & \textbf{0.800 $\pm$ 0.025} & \textbf{0.770 $\pm$ 0.021} & 0.627 $\pm$ 0.000 & 0.999 $\pm$ 0.000 \\
\midrule
\multirow[t]{7}{*}{split-seq} & \href{https://docs.scvi-tools.org/en/stable/user_guide/models/scvi.html}{scVI tools} & $\mathcal{N}(0,I)$ & 0.864 $\pm$ 0.005 & 0.598 $\pm$ 0.008 & 0.705 $\pm$ 0.005 & \textbf{0.861 $\pm$ 0.005} & 0.721 $\pm$ 0.020 & 0.963 $\pm$ 0.007 & 0.956 $\pm$ 0.002 & 0.822 $\pm$ 0.005 & 0.559 $\pm$ 0.003 & 0.317 $\pm$ 0.017 & 0.575 $\pm$ 0.017 & 0.541 $\pm$ 0.004 & 1.000 $\pm$ 0.000 \\
 & \multirow[t]{4}{*}{scVI (our code)} & $\mathcal{N}(0,I)$ & 0.875 $\pm$ 0.006 & 0.612 $\pm$ 0.007 & 0.717 $\pm$ 0.004 & \textbf{0.863 $\pm$ 0.002} & 0.745 $\pm$ 0.023 & \textbf{0.982 $\pm$ 0.003} & \textbf{0.959 $\pm$ 0.002} & 0.825 $\pm$ 0.004 & 0.563 $\pm$ 0.003 & 0.347 $\pm$ 0.018 & 0.604 $\pm$ 0.012 & 0.545 $\pm$ 0.005 & 1.000 $\pm$ 0.000 \\
 &  & VampPrior & \textbf{0.876 $\pm$ 0.005} & 0.638 $\pm$ 0.010 & \textbf{0.733 $\pm$ 0.006} & 0.855 $\pm$ 0.006 & \textbf{0.773 $\pm$ 0.011} & 0.953 $\pm$ 0.011 & 0.953 $\pm$ 0.003 & \textbf{0.848 $\pm$ 0.003} & \textbf{0.577 $\pm$ 0.003} & 0.375 $\pm$ 0.029 & \textbf{0.647 $\pm$ 0.016} & \textbf{0.591 $\pm$ 0.004} & 1.000 $\pm$ 0.000 \\
 &  & GMM & \textbf{0.876 $\pm$ 0.008} & 0.631 $\pm$ 0.006 & 0.729 $\pm$ 0.006 & 0.858 $\pm$ 0.004 & \textbf{0.766 $\pm$ 0.025} & 0.965 $\pm$ 0.009 & 0.952 $\pm$ 0.003 & 0.839 $\pm$ 0.005 & 0.569 $\pm$ 0.005 & 0.375 $\pm$ 0.015 & 0.632 $\pm$ 0.013 & 0.580 $\pm$ 0.005 & 1.000 $\pm$ 0.000 \\
 &  & VMM & \textbf{0.878 $\pm$ 0.004} & 0.639 $\pm$ 0.004 & \textbf{0.734 $\pm$ 0.003} & 0.855 $\pm$ 0.004 & \textbf{0.773 $\pm$ 0.016} & 0.960 $\pm$ 0.005 & 0.953 $\pm$ 0.002 & \textbf{0.846 $\pm$ 0.004} & 0.568 $\pm$ 0.003 & 0.389 $\pm$ 0.015 & \textbf{0.648 $\pm$ 0.007} & \textbf{0.588 $\pm$ 0.004} & 1.000 $\pm$ 0.000 \\
 & Harmony & N/A & 0.790 $\pm$ 0.002 & \textbf{0.650 $\pm$ 0.002} & 0.706 $\pm$ 0.002 & 0.854 $\pm$ 0.004 & 0.606 $\pm$ 0.007 & 0.977 $\pm$ 0.001 & 0.949 $\pm$ 0.001 & 0.563 $\pm$ 0.004 & 0.568 $\pm$ 0.000 & \textbf{0.443 $\pm$ 0.010} & \textbf{0.652 $\pm$ 0.000} & \textbf{0.589 $\pm$ 0.000} & \textbf{1.000 $\pm$ 0.000} \\
 & Scanorama & N/A & 0.552 $\pm$ 0.000 & 0.632 $\pm$ 0.000 & 0.600 $\pm$ 0.000 & 0.604 $\pm$ 0.000 & 0.294 $\pm$ 0.000 & 0.662 $\pm$ 0.000 & 0.858 $\pm$ 0.000 & 0.341 $\pm$ 0.000 & 0.568 $\pm$ 0.000 & 0.395 $\pm$ 0.000 & 0.619 $\pm$ 0.000 & 0.580 $\pm$ 0.000 & 1.000 $\pm$ 0.000 \\
\midrule
\multirow[t]{7}{*}{lung atlas} & \href{https://docs.scvi-tools.org/en/stable/user_guide/models/scvi.html}{scVI tools} & $\mathcal{N}(0,I)$ & \textbf{0.620 $\pm$ 0.006} & 0.608 $\pm$ 0.015 & 0.613 $\pm$ 0.010 & 0.845 $\pm$ 0.018 & 0.354 $\pm$ 0.015 & 0.928 $\pm$ 0.009 & 0.815 $\pm$ 0.012 & 0.157 $\pm$ 0.012 & 0.675 $\pm$ 0.022 & 0.319 $\pm$ 0.035 & 0.528 $\pm$ 0.025 & 0.537 $\pm$ 0.012 & 0.980 $\pm$ 0.007 \\
 & \multirow[t]{4}{*}{scVI (our code)} & $\mathcal{N}(0,I)$ & 0.616 $\pm$ 0.006 & 0.600 $\pm$ 0.008 & 0.607 $\pm$ 0.006 & 0.840 $\pm$ 0.011 & 0.322 $\pm$ 0.011 & \textbf{0.952 $\pm$ 0.009} & 0.821 $\pm$ 0.012 & 0.145 $\pm$ 0.008 & 0.684 $\pm$ 0.022 & 0.289 $\pm$ 0.025 & 0.511 $\pm$ 0.009 & 0.537 $\pm$ 0.007 & 0.980 $\pm$ 0.003 \\
 &  & VampPrior & 0.538 $\pm$ 0.014 & \textbf{0.688 $\pm$ 0.015} & \textbf{0.628 $\pm$ 0.012} & 0.823 $\pm$ 0.004 & \textbf{0.530 $\pm$ 0.012} & 0.449 $\pm$ 0.066 & 0.708 $\pm$ 0.010 & \textbf{0.178 $\pm$ 0.004} & 0.618 $\pm$ 0.035 & \textbf{0.522 $\pm$ 0.041} & 0.679 $\pm$ 0.020 & \textbf{0.633 $\pm$ 0.013} & 0.990 $\pm$ 0.003 \\
 &  & GMM & 0.611 $\pm$ 0.007 & 0.628 $\pm$ 0.019 & \textbf{0.621 $\pm$ 0.013} & \textbf{0.854 $\pm$ 0.010} & 0.360 $\pm$ 0.011 & 0.888 $\pm$ 0.029 & 0.812 $\pm$ 0.009 & 0.141 $\pm$ 0.007 & 0.696 $\pm$ 0.035 & 0.341 $\pm$ 0.052 & 0.551 $\pm$ 0.032 & 0.564 $\pm$ 0.011 & 0.988 $\pm$ 0.002 \\
 &  & VMM & 0.562 $\pm$ 0.012 & 0.674 $\pm$ 0.012 & \textbf{0.629 $\pm$ 0.008} & \textbf{0.860 $\pm$ 0.011} & 0.478 $\pm$ 0.020 & 0.544 $\pm$ 0.059 & 0.767 $\pm$ 0.008 & 0.160 $\pm$ 0.003 & 0.686 $\pm$ 0.031 & 0.461 $\pm$ 0.022 & 0.638 $\pm$ 0.013 & 0.599 $\pm$ 0.009 & 0.988 $\pm$ 0.002 \\
 & Harmony & N/A & 0.553 $\pm$ 0.003 & 0.678 $\pm$ 0.002 & \textbf{0.628 $\pm$ 0.002} & 0.753 $\pm$ 0.004 & \textbf{0.536 $\pm$ 0.010} & 0.459 $\pm$ 0.004 & \textbf{0.881 $\pm$ 0.001} & 0.139 $\pm$ 0.000 & 0.577 $\pm$ 0.003 & \textbf{0.536 $\pm$ 0.009} & \textbf{0.699 $\pm$ 0.005} & 0.579 $\pm$ 0.001 & 0.996 $\pm$ 0.000 \\
 & Scanorama & N/A & 0.328 $\pm$ 0.006 & \textbf{0.694 $\pm$ 0.009} & 0.548 $\pm$ 0.007 & 0.303 $\pm$ 0.021 & 0.337 $\pm$ 0.015 & 0.088 $\pm$ 0.007 & 0.830 $\pm$ 0.002 & 0.081 $\pm$ 0.001 & \textbf{0.744 $\pm$ 0.004} & 0.463 $\pm$ 0.032 & \textbf{0.696 $\pm$ 0.010} & 0.570 $\pm$ 0.003 & \textbf{0.997 $\pm$ 0.000} \\
\bottomrule
\end{tabular}
}
\end{table}

Our scVI implementation is in TensorFlow, whereas the scVI tools version is part of their Python package and based in PyTorch.
It was easier to build scVI with a VMM from scratch than from within the scvi-tools Python package given the complexity of the existing framework.
We faithfully ensure that both implementations are equivalent (e.g.\ network architecture, optimizer hyperparameter, scVI options, etc.) hence the statistically equivalent performance ($p \geq 0.05$) in \cref{tab:sc-integration-performance}.

We use \citet{lopez_deep_2018}\('\)s recommended learning rates for scVI's variational inference steps.
For the VMM, we tuned the batch size and Empirical Bayes learning rate ($\gamma_2$, \cref{alg:aternating-inference}) to optimize \citet{luecken_benchmarking_2022}\('\)s aggregate total score;
per \cref{subsec:aternating-inference}, this amounts to tuning the strength of the hyper-priors.
For all scRNA-seq results, we set $K=100$ as we did for the VAE and allow all models to train for a maximum of 10,000 epochs.
Again, no model comes close to using this many epochs since we halt training early if the validation set's ELBO has not improved for 100 epochs.
As before, we restore model parameters from the best epoch.
Please see our provided code for additional details.

\Cref{tab:sc-integration-performance} reports \citet{luecken_benchmarking_2022}\('\)s aggregate scores from \cref{tab:sc-performance} alongside the scores composing the batch correction and the bio conservation aggregate scores.
They rescale all scores to $[0,1]$ such that larger values denote better performance.
Please refer to their manuscript for further details.

\begin{figure}
    \centering
    \centerline{\includegraphics[width=\columnwidth]{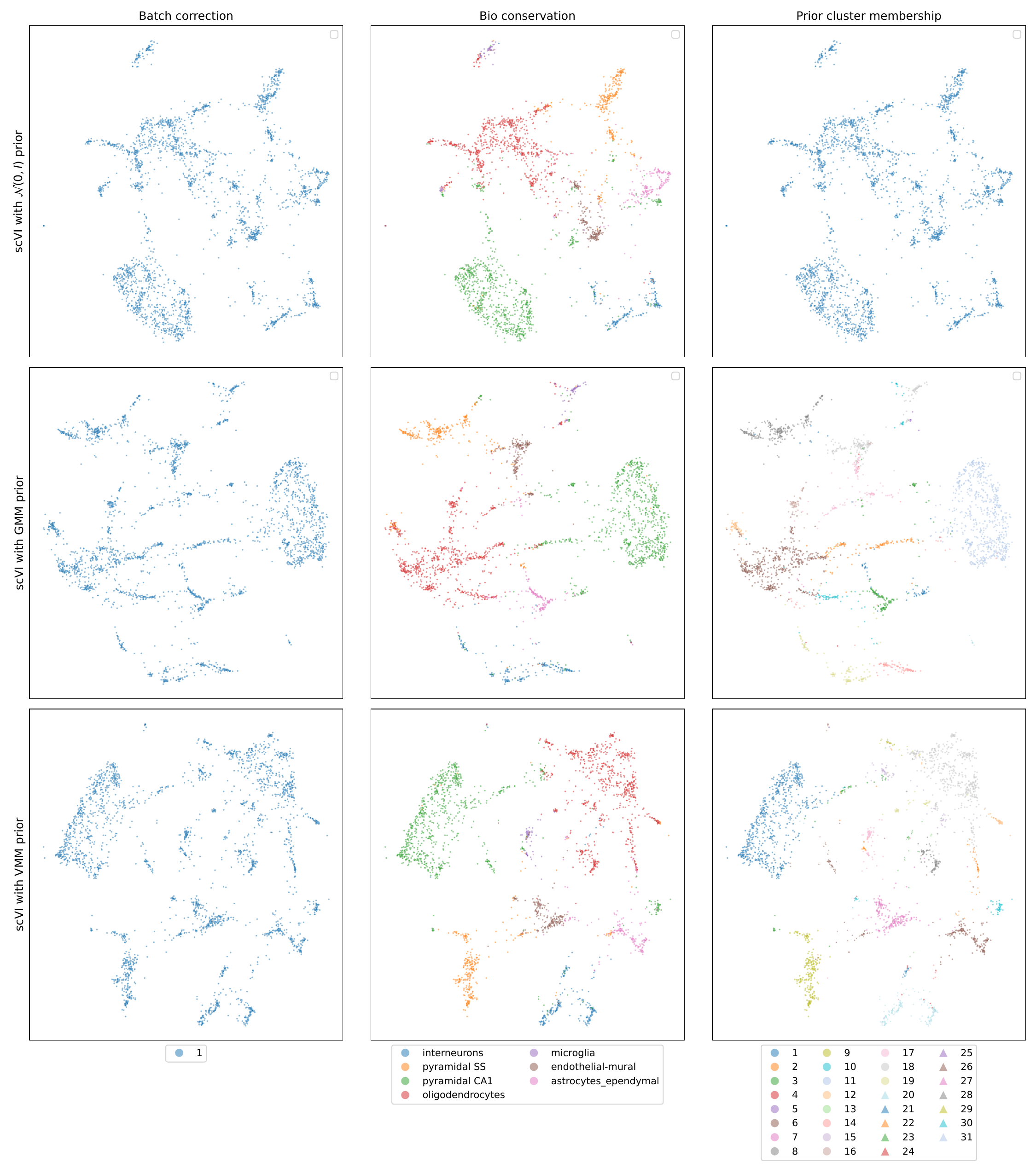}}
    \caption{\mde{cortex}}
\end{figure}

\begin{figure}
    \centering
    \centerline{\includegraphics[width=\columnwidth]{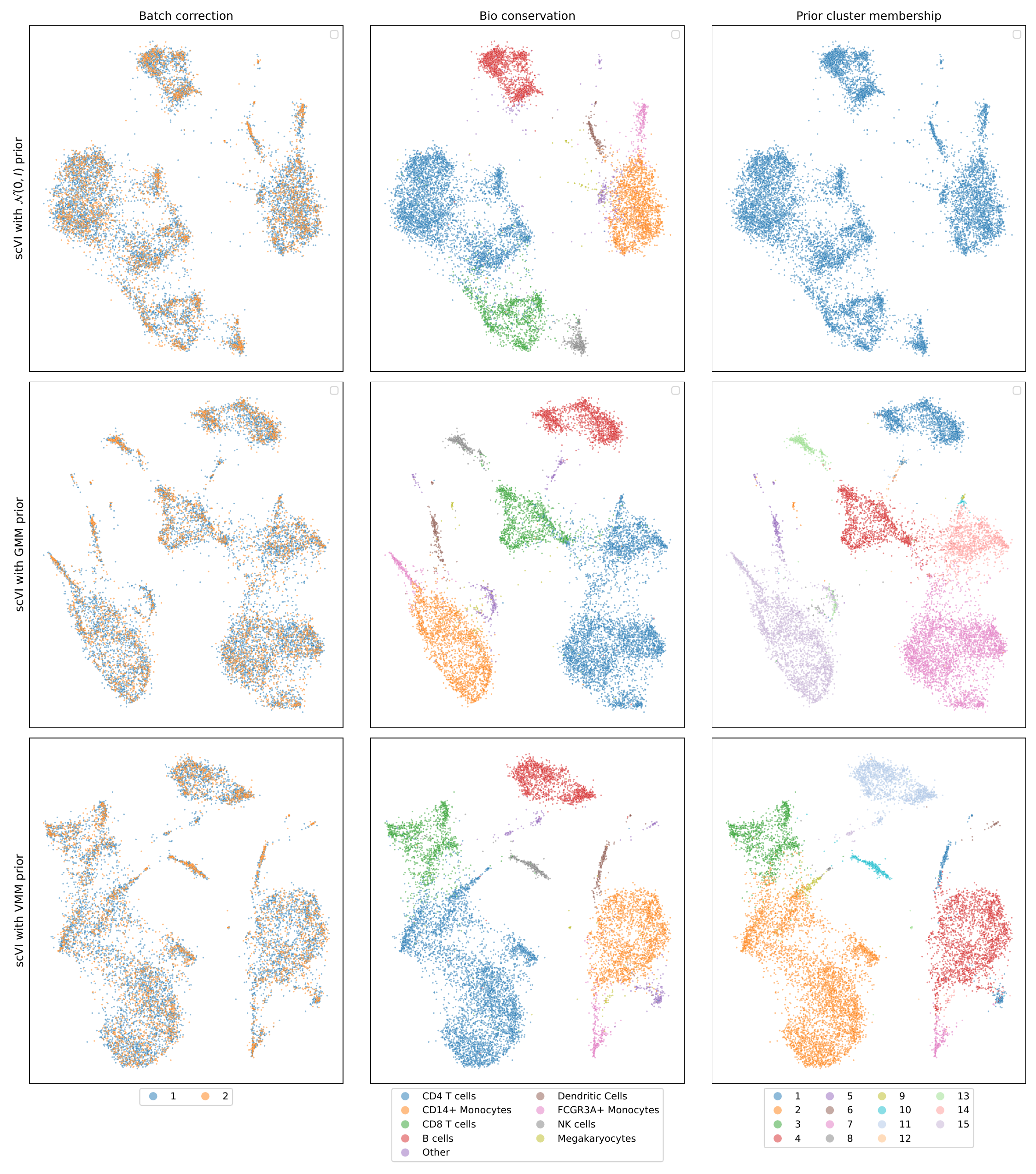}}
    \caption{\mde{PBMC}}
\end{figure}

\begin{figure}
    \centering
    \centerline{\includegraphics[width=\columnwidth]{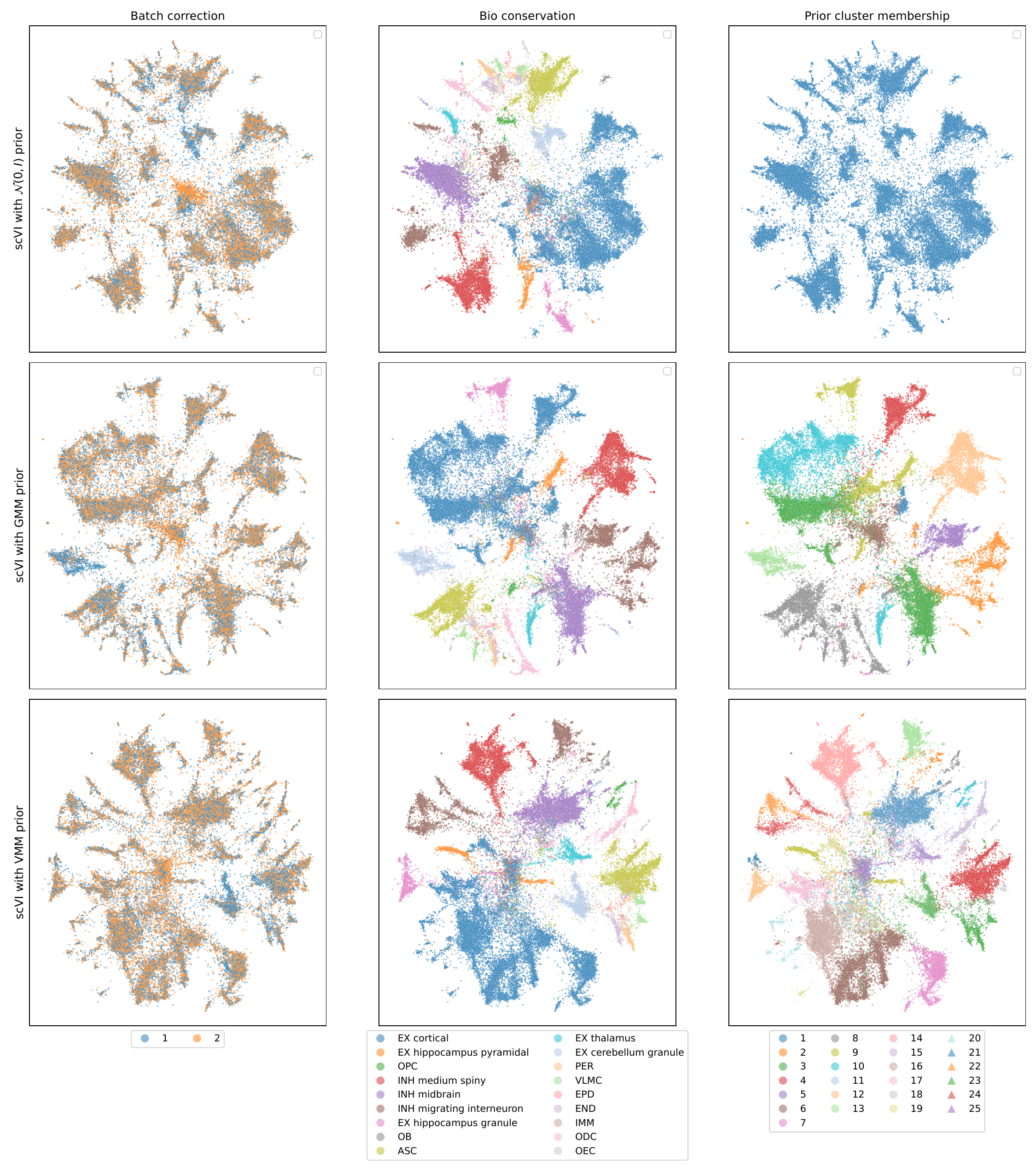}}
    \caption{\mde{split-seq}}
\end{figure}

\end{document}